\documentclass[]{spie}

\usepackage{booktabs}
\usepackage{amsmath,amssymb}
\usepackage{graphicx}
\usepackage[colorlinks=true, allcolors=blue]{hyperref}

\title{Weakly Supervised Spatial Grounding for Discriminative Attention-Based
Ultrasound-Histopathology Alignment in Prostate Cancer Grading}

\author[1,2]{Obed Korshie Dzikunu}
\author[2,3]{Emma Willis}
\author[1,2]{Mohammad Mahdi Abootorabi}
\author[2,3]{Mohamed Harmanani}
\author[2,3]{Zhuoxin Guo}
\author[4]{Ferdinand Luger}
\author[5]{Adam Kinnaird}
\author[6]{Brian Wodlinger}
\author[2,3]{Parvin Mousavi}
\author[1]{Purang Abolmaesumi}

\affil[1]{Department of Electrical and Computer Engineering, University of British Columbia, Vancouver, Canada}
\affil[2]{Vector Institute, Toronto, Canada}
\affil[3]{School of Computing, Queen's University, Kingston, Canada}
\affil[4]{Ordensklinikum Linz, Linz, Austria}
\affil[5]{Department of Surgery, University of Alberta, Edmonton, Canada}
\affil[6]{Exact Imaging, Markham, Canada}

\begin{document}
\maketitle

\authorinfo{Further author information: (Send correspondence to Obed Korshie
Dzikunu)\\
Obed Korshie Dzikunu: E-mail: obed@ece.ubc.ca}

\begin{abstract}
Unpaired cross-modal distillation transfers grade structure from histopathology
into a micro-ultrasound (micro-US) encoder by aligning a pooled needle-region
embedding to a frozen histopathology teacher under grade-group correspondence
alone. A single objective is thereby required to serve two distinct functions: rendering patch features discriminative of tissue state, and selecting which patches enter the pooled representation. We decouple them. Weak spatial supervision derived from percentage involvement, recorded routinely at biopsy, constrains the predicted proportion of malignant tissue within each core, acting on the encoder features independently of the alignment objective. The alignment loss then operates on features that differ across a core, and attention concentrates on a subset of patches rather than remaining near-uniform. On $7,166$ biopsy cores from $811$ patients across seven centers under patient-level 5-fold cross-validation, the method reaches $67.1$ macro AUC and $68.5$ csPCa AUC, against $61.2$ and $52.8$ for the existing unpaired alignment method and $63.1$ and $62.6$ for the strongest unimodal baselines. Ablation against existing attention regularizers designed to prevent attention-uniformity collapse shows that such regularizers do not substitute for label-derived supervision: they constrain the attention distribution, whereas the signal required acts on the features that attention reads.
\end{abstract}

\keywords{Prostate cancer, micro-ultrasound, grade group, cross-modal
distillation, multiple instance learning, weak supervision}

\section{Introduction}

Prostate cancer (PCa) is graded from histopathology, where nuclear architecture,
glandular organization, and stromal disruption are resolved at cellular scale.
But histopathology becomes available only after tissue is extracted and
processed, so it cannot guide where to sample or how to triage during the
procedure itself. Micro-ultrasound (micro-US) is an intraoperative alternative:
at roughly 70 $\mu$m it resolves tissue microstructure finely enough to be
clinically useful, yet the acoustic signal does not directly encode cellular
morphology, and encoders trained on micro-US labels alone struggle to learn the
fine-grained distinctions that drive grading. Even heavily fine-tuned
general-purpose foundation models fall short of specialist performance on this
task, which motivates injecting additional domain knowledge into the micro-US
encoder. ProstNFound and its successor ProstNFound+ do exactly
this\cite{prostnfound,prostnfoundplus}, embedding ultrasound- and PCa-specific
priors into a foundation-model detector, but neither draws on histopathological supervision during training.

Cross-modal distillation offers a complementary source of that knowledge: if the
structure a histopathology encoder has learned can be transferred into the
micro-US encoder during training, the student can learn tissue distinctions it
could not otherwise supervise, while remaining deployable on ultrasound alone at
inference. Where co-registered histopathology is available for the same patient,
this is well established. CorrSigNet\cite{corrsignet} learns MRI signatures
correlated with whole-mount histopathology and discards the histopathology at
inference, and CorrSigNIA\cite{corrsignia} extends this to jointly localize
indolent and aggressive disease. All such methods, however, require image
registration between modalities, which micro-US cannot provide; micro-US frames
and whole-slide histopathology differ too much in resolution and anatomical
coverage for pixel- or patient-level registration at scale. GUIDE-US\cite{guideus} removes this 
requirement by distilling in the unpaired regime, aligning grade-conditioned distributions 
between a frozen histopathology teacher and a micro-US student. It anchors distillation to 
a shared clinical label (ISUP grade) rather than spatial correspondence, and uses attention-based
multiple-instance learning (ABMIL)\cite{ilse2018abmil} to pool needle-region
patch tokens into a single bag embedding before alignment.

\noindent\textbf{Motivation.} In the unpaired alignment regime, the alignment objective is defined on the bag
embedding alone, and is consequently asked to perform two functions at once: to
make the patch features discriminative of tissue state, and to determine which
patches the attention head selects. Neither is supervised directly. Attention 
weights receive gradient only through their effect on the pooled vector, and 
no per-patch target exists. Every token of a core inherits the same core-level 
label, regardless of how much of that core is involved by cancer. Many 
attention distributions therefore produce a pooled embedding that satisfies 
the alignment objective equally well, including distributions concentrated 
on tissue that is not malignant. The generic remedy is to regularize the attention
distribution. ACMIL\cite{acmil} and AEM\cite{aem} were both developed for
whole-slide image classification to counteract attention concentrating on a
small subset of instances, a pattern associated with overfitting. Both act on
the attention distribution rather than on the features that attention reads,
and neither supplies discriminative structure to the representation. In a
needle-biopsy setting the fraction of the core that is
malignant is moreover recorded rather than unknown, so a prior on the shape of
the attention distribution discards information that is already available.

\begin{figure}[t]
\centering
\includegraphics[width=0.82\linewidth]{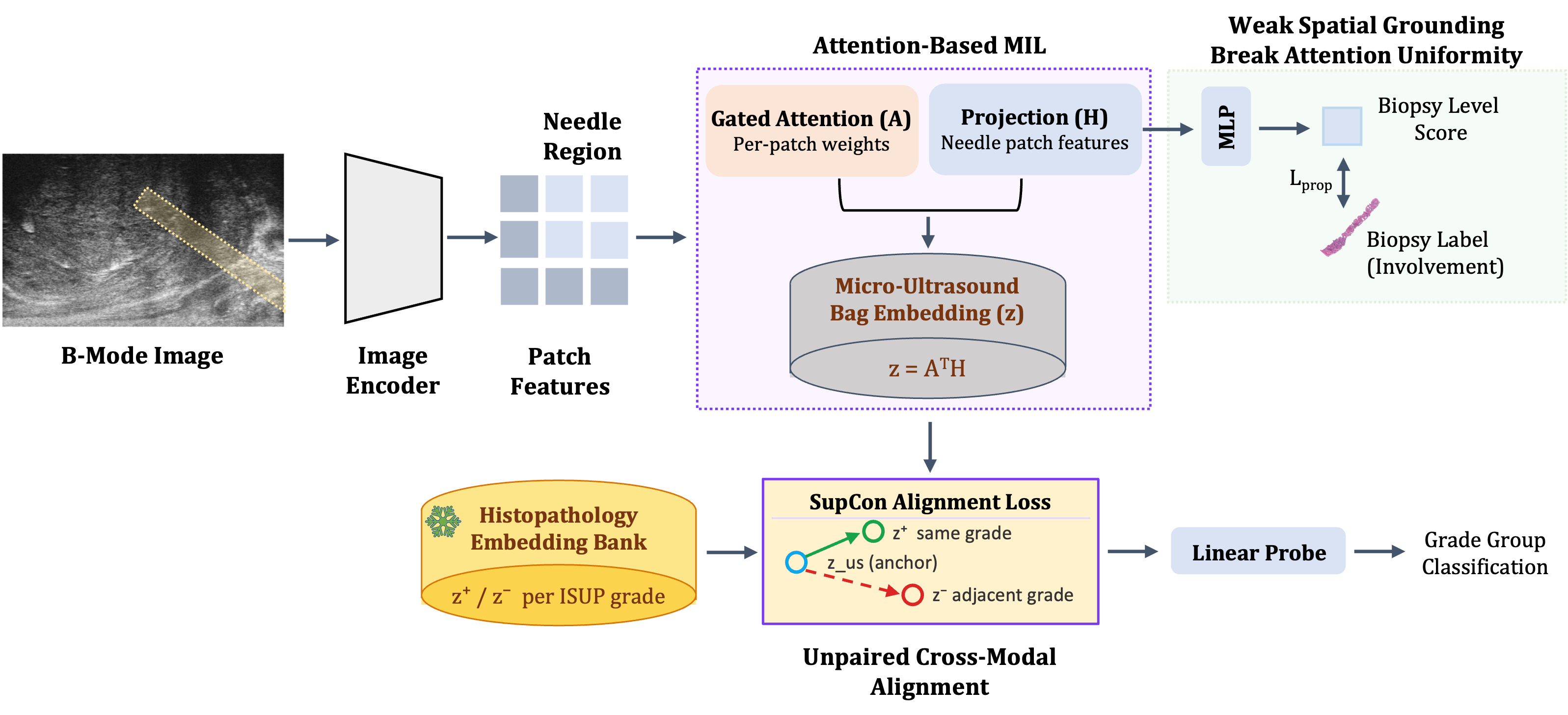}
\caption{AlignUS architecture. B-mode patches are split into needle-region and
background regions using biopsy core geometry. Gated attention ($A$) and
projection ($H$) form a bag embedding ($A^{\top}H$), which is aligned to
histopathology embeddings via $\mathcal{L}_{\text{align}}$ for downstream ISUP
grading. A parallel branch applies an MLP to $H$ to predict per-patch cancer
probability, constrained by $\mathcal{L}_{\text{sg}}$ to match known core
involvement over needle-region patches, breaking attention-uniformity
collapse by injecting a spatially informative gradient.}
\label{fig:overview}
\end{figure}

\noindent\textbf{Contribution.} We propose AlignUS (Fig. \ref{fig:overview}), a micro-US–histopathology alignment framework that supervises these two functions independently. The first, rendering patch features discriminative of tissue state, is supervised by a lightweight head that assigns a cancer score to each needle-region patch, with a proportion-matching loss constraining the mean of those scores to the percentage involvement recorded at biopsy. The second, selecting which patches enter the pooled representation, remains the responsibility of the alignment objective, which now operates over patch features that differ across a core rather than converging toward a common direction. We evaluate against four unimodal baselines and one cross-modal baseline, and ablate against unconstrained attention, mean pooling, and two attention regularizers.

\section{Materials and Methods}

\subsection{Data}

\noindent\textbf{Histopathology (PANDA).} The histopathology dataset is derived
from the Prostate Cancer Grade Assessment (PANDA) challenge\cite{Bulten2022PANDA},
which contains 10,616 whole-slide images (WSIs) of prostate core needle
biopsies. Each slide is annotated with an International Society of Urological
Pathology (ISUP) grade. Preprocessing followed the protocol described in prior
work\cite{Bulten2022PANDA}, excluding slides with erroneous annotations, corrupted scans, or noisy
labels. After filtering, 9,555 WSIs remained, stratified by ISUP grade and
partitioned 80:10:10 into training, validation, and test splits.

\noindent\textbf{Micro-US dataset.} We use a multi-center micro-US dataset
comprising 811 patients across two cohorts: 693 patients imaged between
2013 and 2016 at five institutions using an earlier ExactVu $\mu$US system (Exact
Imaging, Markham, Canada) under untargeted systematic biopsy, and 118 patients
imaged between 2021 and 2024 at two sites using a newer ExactVu system under
combined systematic and targeted biopsy. Sagittal B-mode images were acquired at
28 mm depth and $46.06$ mm width during a 30-second cineloop per core capturing
needle alignment and firing. The final frame immediately prior to needle entry,
co-registered with an annotated needle-trace mask (defined by tip position and
insertion angle at firing), was selected as the representative, artefact-free
image per core. Each core is paired with histopathology-derived cancer diagnosis,
percentage involvement, and ISUP Grade Group (GG $0-5$). Labels are assigned to
the full needle-trace region, introducing minor spatial approximation for
partially involved cores. Most patients contributed $10-12$ systematic cores from
standard sextant regions, yielding 7,166 biopsy cores in total, each with a
corresponding needle-trace binary mask. No clinical metadata (PSA, age, prostate
volume) was used; all predictions are derived from imaging alone.

\noindent\textbf{Preprocessing and evaluation protocol.} B-mode images were
resized from their native $1372 \times 833$ resolution to $512 \times 512$ via
bilinear interpolation and normalized to $[0,1]$. Each US frame was paired with a
histopathology embedding sampled by matching ISUP grade and, loosely, percentage
involvement. We performed label-stratified 5-fold cross-validation with splits
defined at the patient level, so that all cores from a given patient fall
exclusively within either the training or validation subset of a fold.

\subsection{Method}

\noindent\textbf{Patch token extraction and aggregation.} A DINOv3
ViT-L/16 encoder\cite{dinov3} produces a grid of patch tokens per B-mode frame,
of which only those falling within the annotated needle-trace mask are
retained, yielding $N_b$ tokens for core $b$. Because $N_b$ varies across cores,
tokens are padded to the batch maximum and a validity mask $m_b \in \{0,1\}^{N}$
is propagated to every downstream term; all masked quantities below are computed
over valid tokens only. Tokens are linearly projected to a shared dimension to
obtain patch features $H_b \in \mathbb{R}^{N \times d}$. A gated attention
mechanism\cite{ilse2018abmil} takes $H_b$ as input and computes weights $A_b$ with the 
softmax normalized over valid positions only (via $m_b)$; the bag embedding is $z_b = A_b^{\top}H_b$, $\ell_2$-normalized.

\noindent\textbf{Cross-modal alignment.} Following prior unpaired
work\cite{guideus}, $z_b$ is aligned to a frozen histopathology embedding of the
same grade group\cite{UNI2_FM}. We use a supervised contrastive
objective\cite{khosla2020} in place of the triplet loss used previously: for
anchor $i$ with grade $y_i$, all same-grade elements form the positive set
$P(i)$ and all different-grade elements the negatives, and
\begin{equation}
\mathcal{L}_{\text{align}} =
-\frac{1}{|\mathcal{B}|}\sum_{i}\frac{1}{|P(i)|}\sum_{p \in P(i)}
\log\frac{\exp(s_{ip})}
{\sum_{p' \in P(i)}\exp(s_{ip'}) + \sum_{n \in N(i)} w_{in}\exp(s_{in})} ,
\qquad s_{ij} = z_i^{\top}z_j/\tau ,
\label{eq:align}
\end{equation}
where negatives are weighted by grade distance,
$w_{ij} = 1 + \beta\,|y_i - y_j| / d_{\max}$ with $d_{\max}=5$ and $\beta=1$;
$\beta=0$ recovers the standard supervised contrastive loss. The objective is
applied both within micro-US and across the concatenated micro-US and
histopathology batch. Histopathology embeddings are used at training
time only; inference requires micro-US alone.

\noindent\textbf{Involvement-supervised spatial grounding.} Let $\iota_b \in
[0,1]$ be the fraction of core $b$ involved by cancer, taken directly from the
biopsy report. A two-layer head $g$ maps each patch feature to a cancer logit,
and the predicted involved fraction is the masked mean of the corresponding
probabilities,
\begin{equation}
\bar{p}_b \;=\; \frac{1}{\sum_n m_{bn}}\sum_{n} m_{bn}\,
\sigma\!\big(g(H_{bn})\big) ,
\label{eq:prop}
\end{equation}
which is matched to the reported involvement by a binary cross-entropy on
proportions,
\begin{equation}
\mathcal{L}_{\text{sg}} \;=\;
-\frac{1}{\sum_b w_b}\sum_b w_b
\Big[\iota_b \log \bar{p}_b + (1-\iota_b)\log(1-\bar{p}_b)\Big] ,
\label{eq:sg}
\end{equation}
where $w_b$ is a per-core weight that offsets the benign majority and the grade
imbalance. The total objective is $\mathcal{L} = \mathcal{L}_{\text{align}} +
\lambda_{\text{sg}}\mathcal{L}_{\text{sg}}$. Two properties give the objective
its intended division of labour. First, the gradient of
Eq. \eqref{eq:sg} reaches the shared patch features $H_b$ and not the attention
weights $A_b$: the term supplies discriminative structure to the representation
without prescribing how that structure should be pooled, in contrast to ACMIL and
AEM, which constrain the attention distribution directly. Second,
Eq. \eqref{eq:sg} constrains only the mean of the patch scores and does not
penalize a uniform solution, so within-core heterogeneity arises as a consequence
of satisfying the proportion constraint rather than as a directly optimized
target. Attention selection remains governed by
$\mathcal{L}_{\text{align}}$, which now operates on features that are weakly supervised to differ across patches in tissue state.

\noindent\textbf{Implementation.} $\lambda_{\text{sg}}$ is empirically set to 1 and $w_b$ is selected on the validation split. All ablation configurations
(Table \ref{tab:ablation}) share the same backbone, pooling
head, batch construction, and optimizer schedule; the unimodal baselines in
Table \ref{tab:main} differ in backbone by construction and are trained with
their published configurations.

\noindent\textbf{Evaluation.} No loss is applied to the classification head; all
reported numbers come from a multinomial logistic regression probe fit on the
frozen bag embedding, applied identically to every method. We report
one-vs-rest AUC per grade group, macro AUC, and binary AUC for clinically
significant cancer (csPCa, GG$\geq$2).

\section{Results}

\begin{table}[t]
\caption{Per-grade one-vs-rest AUC ($\times100$, mean$\pm$std over 5 patient-level
folds). Best per column in bold. Cross-modal methods use histopathology at
training time only.}
\label{tab:main}
\centering
\small
\begin{tabular}{lcccccccc}
\toprule
\textbf{Method} & \textbf{GG0} & \textbf{GG1} & \textbf{GG2} & \textbf{GG3} &
\textbf{GG4} & \textbf{GG5} & \textbf{Macro} & \textbf{csPCa}\\
\midrule
\multicolumn{9}{l}{\textit{Unimodal}} \\
DINO-FT        & 64.5$\pm$2.0 & 69.2$\pm$4.6 & \textbf{59.2$\pm$3.2} & 57.9$\pm$3.5 & 65.0$\pm$7.2 & 60.8$\pm$12.6 & 62.3$\pm$3.3 & 62.6$\pm$2.5 \\
MicroSegNet-FT & 68.1$\pm$4.4 & 90.8$\pm$1.4 & 51.6$\pm$4.7 & 50.1$\pm$5.8 & 62.4$\pm$12.9 & 55.6$\pm$15.7 & 63.1$\pm$4.1 & 57.2$\pm$8.1 \\
MedSAM-FT      & 66.9$\pm$4.5 & \textbf{91.4$\pm$1.3} & 52.2$\pm$3.1 & 50.4$\pm$7.2 & 59.4$\pm$11.2 & 54.0$\pm$14.3 & 62.3$\pm$3.8 & 55.9$\pm$5.9 \\
ProstNFound+   & 61.1$\pm$6.7 & 90.6$\pm$2.0 & 51.4$\pm$5.0 & 49.2$\pm$6.5 & 59.5$\pm$1.2 & 61.1$\pm$8.1 & 62.2$\pm$2.2 & 55.7$\pm$7.1 \\
\midrule
\multicolumn{9}{l}{\textit{Cross-modal alignment}} \\
GUIDE-US       & 65.5$\pm$4.8 & 91.2$\pm$2.1 & 50.7$\pm$1.6 & 49.0$\pm$7.2 & 57.3$\pm$11.2 & 53.4$\pm$12.2 & 61.2$\pm$3.8 & 52.8$\pm$6.7 \\
\textbf{Ours}  & \textbf{69.9$\pm$1.4} & 83.7$\pm$6.7 & 59.0$\pm$6.2 & \textbf{59.8$\pm$5.4} & \textbf{67.5$\pm$8.2} & \textbf{62.7$\pm$10.5} & \textbf{67.1$\pm$4.1} & \textbf{68.5$\pm$2.7} \\
\bottomrule
\end{tabular}
\end{table}

\begin{table}[t]
\caption{Pooling and supervision ablation (top); the bottom two rows test
whether the gain from $\mathcal{L}_{\text{sg}}$ transfers to a second,
unrelated alignment objective (triplet), rather than being specific to the
contrastive loss used elsewhere. All rows share the DINOv3 backbone, batch
construction, and evaluation protocol; unless noted, the alignment objective
is contrastive (Eq. \eqref{eq:align}).}
\label{tab:ablation}
\centering
\small
\begin{tabular}{lcc}
\toprule
\textbf{Configuration} & \textbf{Macro AUC} & \textbf{csPCa AUC}\\
\midrule
Micro-US only (no alignment)                       & 62.3$\pm$3.3 & 62.6$\pm$2.5 \\
Mean pooling, w/o $\mathcal{L}_{\text{sg}}$        & 63.4$\pm$1.3 & 65.2$\pm$3.4 \\
ABMIL, w/o $\mathcal{L}_{\text{sg}}$               & 65.0$\pm$2.9 & 66.9$\pm$3.1 \\
ABMIL + AEM                                        & 65.8$\pm$3.3 & 66.6$\pm$2.2 \\
ABMIL + ACMIL                                      & 65.8$\pm$3.9 & 66.4$\pm$2.6 \\
\textbf{ABMIL + $\mathcal{L}_{\text{sg}}$ (ours)}  & \textbf{67.1$\pm$4.1} & \textbf{68.5$\pm$2.7} \\
\midrule
ABMIL, triplet alignment, w/o $\mathcal{L}_{\text{sg}}$ & 63.9$\pm$4.2 & 54.3$\pm$7.3 \\
ABMIL, triplet alignment $+\ \mathcal{L}_{\text{sg}}$   & 64.2$\pm$5.4 & 63.2$\pm$2.4 \\
\bottomrule
\end{tabular}
\end{table}

\noindent\textbf{Comparison to prior work.} Table \ref{tab:main} shows performance across unimodal and cross-modal frameworks. GUIDE-US attains the lowest macro AUC in the comparison ($61.2$), below
every unimodal baseline, and is at or near chance on clinically significant
cancer (csPCa $52.8$) and on the interior grades (GG3 49.0, GG5 53.4). The proposed
method reaches 67.1 macro AUC and 68.5 csPCa AUC, exceeding GUIDE-US by 5.9 and
15.7 points respectively and the strongest unimodal baseline by 4.0 and 5.9. It
is best on GG0, GG3, GG4, GG5, macro, and csPCa; GG2 is matched by DINO-FT (59.2
versus 59.0), within one standard deviation.

The three foundation-model fine-tunes that lead on GG1 (MicroSegNet-FT, MedSAM-FT,
ProstNFound+, all above 90) fall to $55.7 - 57.2$ on csPCa, and GUIDE-US to $52.8$.
This gap suggests that high GG1 performance under these methods may reflect separation of low-grade tissue from the remainder, rather than a representation that supports the clinically consequential distinction, though label noise near the GG1/GG2 boundary and prevalence effects on the csPCa estimate could also contribute to the drop. he proposed method reverses this trade-off: GG1 AUC drops to 83.7, well below the foundation-model fine-tunes' ($>90$), while csPCa AUC rises by $11 – 16$ points over the same baselines. This is consistent with the behavior of $\mathcal{L}_{\text{sg}}$ which assigns GG1 cores a non-zero involvement target rather than treating them as negative. As a result, low-grade tissue is pulled toward the malignant end of the patch-score range, likely blurring its separability from higher-grade tissue and accounting for the drop in GG1 AUC.

\noindent\textbf{Ablation.} Table \ref{tab:ablation} isolates pooling and its
supervision. Attention pooling accounts for $1.6$ macro AUC points over mean
pooling ($65.0$ versus $63.4$), and adding the cross-modal term to micro-US-only
training accounts for a further $2.7$ ($62.3$ to $65.0$). Neither generic attention
regularizer improves on unconstrained attention: AEM and ACMIL both reach $65.8$
macro against $65.0$, and both reduce csPCa AUC ($66.6$ and $66.4$
against $66.9$). Supervising the patch features with $\mathcal{L}_{\text{sg}}$
instead yields $67.1$ macro and $68.5$ csPCa. The results thus suggest that constraining the attention distribution does not substitute for supplying discriminative structure to the features it reads.

\begin{table}[t]
\caption{Representation-level mechanism check, averaged over the evaluation
set. Cosine dispersion is $1-$mean pairwise cosine similarity between valid
patch features within a core (post-projection $H$); higher values indicate
more within-core heterogeneity. Attention entropy is $H/\ln N_b$, the Shannon
entropy of the attention distribution over a core's $N_b$ valid patches,
normalized per core by $\ln N_b$ rather than a batch-wide maximum; values
near 1 indicate uniform attention, values near 0 indicate concentration on a
few patches.}
\label{tab:mechanism}
\centering
\small
\begin{tabular}{lcc}
\toprule
\textbf{Configuration} & \textbf{Cosine dispersion} & \textbf{Attention entropy}\\
\midrule
ABMIL, w/o $\mathcal{L}_{\text{sg}}$              & 0.07 & 0.93 \\
ABMIL + AEM                                       & 0.06 & 0.99 \\
ABMIL + ACMIL                                     & 0.07 & 0.97 \\
\textbf{ABMIL + $\mathcal{L}_{\text{sg}}$ (ours)} & \textbf{0.12} & \textbf{0.30} \\
\bottomrule
\end{tabular}
\end{table}

\noindent\textbf{Mechanism validation.} Table \ref{tab:mechanism} tests the two
causal steps claimed in Section 2.2. Because ABMIL without $\mathcal{L}_{\text{sg}}$
shares every component with our full method except $\mathcal{L}_{\text{sg}}$
itself, the difference between these two rows isolates its effect. Without
$\mathcal{L}_{\text{sg}}$, attention stays close to uniform (0.93), similar to
ACMIL (0.97) and AEM (0.99); the alignment gradient alone gives attention no
basis to discriminate. Adding $\mathcal{L}_{\text{sg}}$ increases within-core
patch-feature dispersion by roughly 70\% (0.07 to 0.12) and collapses
attention entropy to 0.30, suggesting that the induced feature heterogeneity
is what the alignment gradient subsequently reads out as concentrated
attention. ACMIL and AEM, by contrast, hold entropy at or above the
baseline. That is expected as they are both designed to resist attention concentration
as a generic defence against overfitting in whole-slide classification, the opposite of what a needle-biopsy core with a genuinely partial,
recorded involvement fraction calls for.

\noindent\textbf{Interaction between the two terms.} To test whether the gain depends specifically on the contrastive alignment objective used elsewhere in the paper, the bottom rows of the Table \ref{tab:ablation} pair $\mathcal{L}_{\text{sg}}$ with an
with a different, alignment objective (triplet) and compare the results. On csPCa,
$\mathcal{L}_{\text{sg}}$ improves both objectives ($+8.9$ under triplet, $+1.6$
under contrastive), whereas on macro AUC the gain is negligible under triplet
($+0.3$) and substantial only under contrastive ($+2.1$). Detection therefore
benefits under either alignment objective, while the gain in grade
discrimination is seen to be better with the contrastive loss, consistent with
$\mathcal{L}_{\text{sg}}$ supplying structure to the representation while the
alignment term governs how much of it reaches the pooled embedding.

\section{Discussion and Conclusion}

In unpaired distillation, the alignment objective typically does two jobs at once: making patch features discriminative and selecting among them. Separating these, supervising the first with percentage involvement already recorded at biopsy, adds no annotation cost and leaves attention selection to the alignment term alone. Because inference uses micro-US only, the method remains deployable intraoperatively, when histopathological grading is unavailable. The ablations localize the gain to supervision of the patch features rather than the attention distribution, and the crossing experiment confirms the two terms play distinct, separable roles.

Patch-level annotations however are unavailable in this cohort, so while Table \ref{tab:mechanism} shows gains consistent with $\mathcal{L}_{\text{sg}}$ inducing feature level discrimination, we cannot verify that the patches this heterogeneity privilages correspond to the true spatial distribution of malignant tissue.

\end{document}